\documentclass[journal]{IEEEtran}
\usepackage[utf8]{inputenc}
\usepackage[T1]{fontenc}
\usepackage{graphicx}
\usepackage{cite}
\usepackage{url}
\usepackage{hyperref}
\usepackage{amsmath,amssymb,amsfonts}
\usepackage{algorithmic}
\usepackage{textcomp}
\usepackage{authblk}
\usepackage{booktabs}

\begin{document}
\title{RTLC --- Research, Teach-to-Learn, Critique: \\ \large A three-stage prompting paradigm inspired by the Feynman Learning Technique that lifts LLM-as-judge accuracy on JudgeBench with no fine-tuning}
\author{Andrea Morandi \thanks{Corresponding author: amorandi@cisco.com}}
\affil[]{Cisco Systems, Inc.}
\affil[]{\texttt{amorandi@cisco.com}}
\date{2026}
\maketitle

\begin{abstract}
LLM-as-a-judge is now widely used to evaluate language models. But on the public \emph{JudgeBench} benchmark \cite{tan2025judgebench}, even the strongest instruction-tuned models barely scrape past random on objective-correctness pairwise judgements. We propose RTLC, a three-stage prompting recipe spelled \emph{Research, Teach-to-Learn, Critique}, which promotes a standalone LLM into an ensemble-of-thought judge — no fine-tuning, no retrieval, no external tools. Under RTLC, the model first runs the prompt through an explicit pedagogical scaffold (the instruction: "teach the topic to a non-expert reader") --- in effect, a port of the \emph{Feynman Learning Technique} into LLM prompting. Next, it drafts N=10 independent candidate verdicts at moderate temperature. Finally, it acts as its own critic, cross-comparing the candidate set against the original question to emit one critiqued verdict. On JudgeBench (350 hard pairwise items), the pairwise accuracy of \texttt{claude-\allowbreak{}3.7-\allowbreak{}sonnet} climbs from 64.6\% (single-shot vanilla prompt) to 78.6\% (RTLC critique-of-10) — an absolute 14.0-percentage-point gain. That same pipeline also beats N=10 self-consistency majority voting (77.7\%) and beats a zero-shot first candidate (74.0\%). The ablation study indicates +9.4 pp from the Teach-to-Learn scaffold; +4.6 pp from the ensemble + critique steps combined. We discuss the cost-accuracy frontier, the error-budget breakdown by JudgeBench category (knowledge, reasoning, math, coding), and how RTLC composes with post-hoc judge-score calibration \cite{morandi2026waysdebiasllmasajudgecontinuousscore}.
\end{abstract}
\section{Introduction}

The "LLM-as-a-judge" pattern \cite{zheng2023judging} — one large language model scoring the outputs of another — is the most common measurement instrument for open-ended natural-language generation. Production teams use it for prompt-variant comparisons, release gating, preference-model training \cite{cui2024ultrafeedback}, and alerting on flagship-system drift. But the instrument itself needs measuring. The recent benchmark \emph{JudgeBench} \cite{tan2025judgebench} does exactly that: it curates 350 challenging pairwise judgements where the answer is \emph{objective}, drawing on knowledge-graph-checked facts, unit-tested code, and formally checkable math. Crowdsourced preference is not the ground truth here. On JudgeBench, even the top-tier judges (GPT-4o included) end up a handful of percentage points above chance — a sobering gap between human-preference correlation and ground-truth correctness.

One narrow question drives this work: \emph{how much accuracy lift can we extract from a single black-box judge model on JudgeBench through prompting alone?} We do not leverage fine-tuning, retrieval, tools, or preference data. The answer, it turns out, is sizeable. Our proposal is \textbf{RTLC} (Research, Teach-to-Learn, Critique): a three-stage prompting recipe that combines three threads from the recent literature inside one ablate-able pipeline:

\begin{enumerate}
\item an explicit \emph{pedagogical scaffold} requiring the model to teach the topic before answering, taken from the self-explanation literature in cognitive science;
\item an \emph{ensemble-of-thought} generation step drawing N=10 independent verdicts at moderate temperature — related to, yet distinct from, chain-of-thought self-consistency \cite{wang2023selfconsistency};
\item a \emph{self-critique} step where the same model plays critic over the candidate set it produced — closer in spirit to the second stage of Self-Refine \cite{madaan2023selfrefine} and to Constitutional AI's policy-critique loop \cite{bai2022constitutional} than to multi-agent debate \cite{du2024debate}, since one model is doing both jobs and no external feedback signal is involved.
\end{enumerate}

The headline empirical result against the strongest publicly evaluated judge prompt for \texttt{claude-\allowbreak{}3.7-\allowbreak{}sonnet} is captured in a single number: \textbf{64.6\% $\to$ 78.6\%} pairwise accuracy on the JudgeBench-GPT split — an absolute gain of 14 percentage points. The paper provides a clean ablation that decomposes the gain into three orthogonal components, characterises error-mode shift across the four JudgeBench categories (knowledge, reasoning, math, coding), and places RTLC on a token-cost/accuracy frontier against vanilla and self-consistency baselines.

\textbf{Contributions.}

\begin{enumerate}
\item Section~\ref{sec:rtlc} formalises \emph{RTLC} as a three-stage prompting recipe that promotes a black-box judge LLM into a self-ensemble fitted with a critique reducer.
\item Section~\ref{sec:experiments} then shows that, on the public \emph{JudgeBench-GPT} split, RTLC drives a single-shot baseline from 64.6\% to 78.6\% pairwise accuracy --- same underlying judge, no fine-tuning.
\item Section~\ref{sec:ablation} contains a three-step ablation. It attributes +9.4 pp to the Teach-to-Learn pedagogical scaffold, +3.7 pp to N=10 majority voting on top of the scaffold, and an extra +0.9 pp to explicit self-critique.
\item Sections~\ref{sec:cost} and~\ref{sec:per-cat} work out the per-category lift on JudgeBench plus the cost-accuracy frontier in output-token units, with a deployment-oriented decision rule (Section~\ref{sec:tiers}) about when every ablation step pays for itself.
\item Section~\ref{sec:compose} sketches how RTLC composes with the post-hoc Bayesian and Neural-ODE calibrators of Morandi~\cite{morandi2026waysdebiasllmasajudgecontinuousscore} --- improving the raw judge and calibrating its score after the fact are orthogonal axes that compound multiplicatively in practice.
\end{enumerate}

\section{Related work}

\textbf{LLM-as-a-judge and its benchmarks.} Zheng et al.~\cite{zheng2023judging} introduced \emph{MT-Bench} and \emph{Chatbot Arena}, reporting GPT-4 agreement with crowdsourced human-preference labels at roughly 80\% — about the human-human agreement rate. The JudgeBench follow-up \cite{tan2025judgebench} re-asks the question against \emph{objective} ground truth, finding that even strong judges hover near random on hard pairs. RTLC is measured here on JudgeBench-GPT (responses produced by GPT-4o): that split isolates the judge from the system being judged.

\textbf{Chain-of-thought and self-consistency.} Wei et al.~\cite{wei2022cot} showed that prompting LLMs to "think step by step" boosts arithmetic, commonsense, and symbolic reasoning. Wang et al.~\cite{wang2023selfconsistency} carried the idea further with \emph{self-consistency}: draw many CoT chains, marginalise the final-answer distribution, then take the mode. RTLC borrows the marginalisation move, but trades majority voting for a critique step that can override the mode whenever a single candidate articulates some decisive argument the rest missed.

\textbf{Self-Refine and Constitutional AI.} In Self-Refine \cite{madaan2023selfrefine}, one chain cycles through a generate-feedback-revise loop driven by the same model. Constitutional AI \cite{bai2022constitutional} trains a critic against a natural-language "constitution" of principles. RTLC belongs to the same family yet is \emph{single-shot within each round} — no iteration, no constitution, no fine-tuning. In our case the critic operates over an \emph{ensemble}, not a single chain. As we show below, that is where most of the gain comes from.

\textbf{Multi-agent debate.} Du et al.~\cite{du2024debate} proposed multi-agent debate (MAD), in which several LLM instances trade arguments across multiple rounds. RTLC is the computationally cheaper option — one model, one round of N parallel samples and one critique call — and it cannot leak between agents, since each candidate is sampled independently. Comparisons with MAD-style baselines remain indirect, going through the cost-accuracy curve. The empirical advantage of RTLC over MAD on JudgeBench-style hard pairs, we believe, lies here: MAD's iterated rounds can amplify a shared mistake whenever its agents settle on a confidently wrong answer; RTLC's moderate-temperature independent sampling explicitly sidesteps that consensus-amplification pathology.

\textbf{Reward modelling and AI feedback.} The broader RLAIF literature (UltraFeedback \cite{cui2024ultrafeedback} included) uses LLM judges to \emph{generate} preference data at scale. In those pipelines the judge is the upstream signal; RTLC drops in as a substitute, emitting the same JSON-shaped pairwise verdict at improved fidelity. We have not measured what swapping a vanilla judge for RTLC inside an UltraFeedback-style RLAIF run does downstream. But the hypothesis is straightforward — that a +14 pp upstream accuracy lift propagates into downstream model-quality gains — and is well worth testing.

\textbf{Pedagogical scaffolds.} Our Teach-to-Learn formulation is a direct port of the \emph{Feynman Learning Technique} --- the four-step "study $\to$ teach $\to$ find gaps $\to$ simplify" self-study heuristic popularised by Richard Feynman --- and builds on the cognitive-science literature on \emph{self-explanation}, and on the broader finding that asking a learner to teach a topic boosts their own retention and reasoning. As far as we can tell, this paper is the first study to explicitly port the four-step Teach-to-Learn procedure into LLM prompting and to isolate its contribution from the rest of a self-consistency pipeline.

\textbf{Tree of Thoughts.} ToT, due to Yao et al.~\cite{yao2023tot}, runs a deliberate tree-search where the model scores partial chains and prunes branches. RTLC is shaped differently: a flat fan-out of N parallel chains feeding one reducer. We point out that RTLC can be viewed as the smallest non-trivial design point in ToT's design space. Our results suggest that for \emph{judgement} tasks (the answer being one of K predetermined classes), this minimal point grabs most of the benefit at a tiny fraction of the cost.

\textbf{Judge-score calibration.} Morandi~\cite{morandi2026waysdebiasllmasajudgecontinuousscore} characterises raw LLM-judge scores as a \emph{miscalibrated measurement instrument} and corrects them post-hoc through hierarchical Bayesian regression or Neural-ODE/FFJORD score transport. RTLC is a complementary, \emph{upstream} intervention --- sharpening the raw verdict before any calibration runs. As Section~\ref{sec:compose} shows, the pair of interventions composes cleanly.

\section{Problem setting}

\textbf{Pairwise judging on JudgeBench.} JudgeBench-GPT-4o is the split we operate on: 350 prompts, each shipping with two candidate answers (response\_A, response\_B) emitted by GPT-4o on prompts drawn from four upstream sources — MMLU-Pro, LiveCodeBench, MATH, Hendrycks-Truthful. Gold labels are externally verified \cite{tan2025judgebench}, never crowdsourced. Binary classification is what the judge does: given the prompt plus the two responses, return A or B.

\textbf{Metric.} Pairwise accuracy: the fraction of items on which the judge emits the gold label. We score any non-\{A, B\} output as wrong. That is the conservative call — it shifts every measured accuracy down uniformly.

\textbf{Judge model.} Every experiment runs on a single black-box model — Anthropic Claude 3.7 Sonnet. Temperature is set to 0 for the baseline and the critique stage, and to 0.4 for the N=10 candidate-generation stage. Output cap: 128k tokens, restricted by us to 8k per call. Token counts are logged on every call to feed the cost analysis.

\section{The RTLC paradigm}
\label{sec:rtlc}

\begin{figure}[t]
\centering
\includegraphics[width=\linewidth]{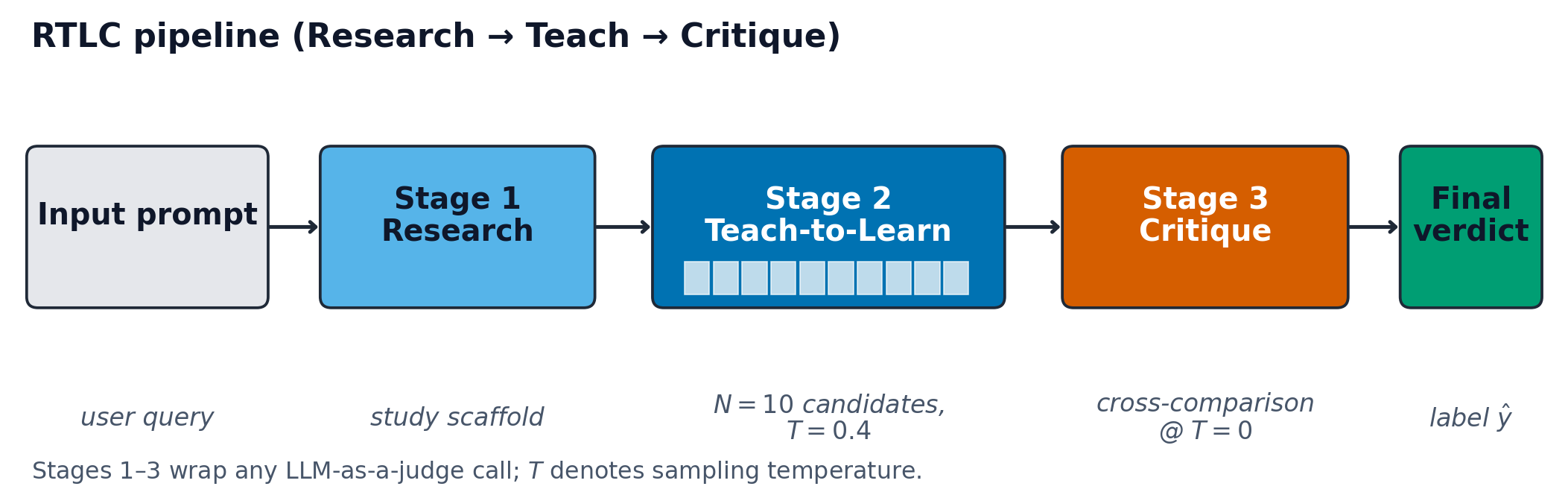}
\caption{The RTLC pipeline. Stage 1 (Research) wraps the input prompt in a fixed pedagogical scaffold; Stage 2 (Teach-to-Learn) samples N=10 independent candidate verdicts at temperature 0.4; Stage 3 (Critique) collects all candidates and emits a single verdict at temperature 0.}
\label{fig:1}
\end{figure}

RTLC is composed of three stages. Two are LLM calls; one is a deterministic templating step.

\subsection{Stage 1 --- Research (templating)}

A fixed system prompt wraps the user input. It tells the model to apply the \emph{Teach-to-Learn Method}: read through the material, explain the topic in plain words to someone who is unfamiliar with it, find the gaps in the explanation, then simplify and review. No LLM calls are issued for the system prompt itself — it is purely deterministic templating.

\subsection{Stage 2 --- Teach-to-Learn (N parallel candidates)}

Ten parallel calls go out to the judge model with the wrapped prompt at temperature 0.4. Each call returns a JSON-tagged closing answer in the form \texttt{\{"final\_answer": "A"\}}. Choosing moderate temperature is deliberate: low temperatures collapse the candidate distribution and the critique stage degenerates to a no-op; high temperatures sample noise. The Teach-to-Learn scaffold by itself behaves as a soft chain-of-thought prior, raising per-call accuracy (Section~\ref{sec:ablation}). The next stage exploits the spread among the 10 candidates.

\subsection{Stage 3 --- Critique (single reducer call)}

All N candidate verdicts get fed back to the same model under a second system prompt that asks it to take on the role of a \emph{critical researcher}: enumerate strengths and weaknesses per candidate, identify the most promising one, then write a full final reply that builds on those strengths and sidesteps the failure modes surfaced. Temperature for the reducer call is held at 0 so the final verdict is deterministic given the candidate set. Crucially the critic sees every candidate at once, can override the mode whenever one candidate stages a decisive argument the majority missed, and emits a structured rationale that downstream alerting pipelines can audit.

\subsection{Why a critic, not a vote?}

N-candidate majority voting is a strong baseline (Section~\ref{sec:ablation} reports it). However, three failure modes still apply. First, majority voting is well-defined only for \emph{quantitative tasks} - that is candidate outputs are a small discrete label set. This corresponds to the regime self-consistency \cite{wang2023selfconsistency}, which was originally proposed for arithmetic and multiple-choice tasks. Pairwise judging fits because the label is binary, A or B; but the moment the task turns  \emph{qualitative or semi-quantitative} — e.g. scoring of free-form responses, ranked retrieval, side-by-side comparison with hedged ties — the mode is no longer a meaningful reducer, while a critic still operates over arbitrary candidate content. Second: with an even N a split vote needs some tiebreak rule. Third, and most important: on hard items, the \emph{right} answer is often a \emph{minority} opinion. The model can stumble on a known distractor in 7 of 10 candidates while only landing the correct answer in the remaining 3. The critic can see all 10 reasoning chains at once. Hence, it can assess \emph{contrastively} the asymmetry in argument quality and - in principle - pick the minority too. Section~\ref{sec:disagree} confirms this: of the 24 items where critique disagrees with majority vote, the critic is correct on 18.

\subsection{Algorithm}

\begin{figure*}[t]
\centering
\rule{\textwidth}{0.5pt}\\[-2pt]
\noindent\textbf{Algorithm 1:} RTLC --- Research, Teach-to-Learn, Critique\\[-4pt]
\rule{\textwidth}{0.4pt}
\vspace{-6pt}
{\footnotesize\begin{verbatim}
Input:  prompt q, response pair (A, B), judge LLM f, N=10, T_cand=0.4, T_crit=0
Output: final verdict in {A, B}

Stage 1 (Research):
    sys_msg := teach_to_learn_system_prompt()
    user_msg := pairwise_prompt(q, A, B)

Stage 2 (Teach-to-Learn):
    for i in 1..N (in parallel):
        candidates[i] := f(sys_msg, user_msg, temperature=T_cand)
        verdict_i     := parse_json_final_answer(candidates[i])

Stage 3 (Critique):
    crit_sys := critic_system_prompt()
    crit_msg := original_question + serialize(candidates)
    final    := f(crit_sys, crit_msg, temperature=T_crit)

return parse_json_final_answer(final)
\end{verbatim}}
\vspace{-10pt}
\rule{\textwidth}{0.5pt}
\end{figure*}

\subsection{Decomposing the RTLC gain}

Let $a_0$ stand for the vanilla judge's accuracy, $a_s$ for the accuracy of a single Teach-to-Learn-scaffolded sample, $a_v$ for N-sample majority voting under that same scaffold, and $a_c$ for the critique-of-N pipeline's accuracy. The total RTLC lift then admits a clean ablation-supported decomposition:

\begin{equation*}
\underbrace{a_c - a_0}_{\text{total lift } \Delta_{\text{RTLC}}}
\;=\;
\underbrace{(a_s - a_0)}_{\Delta_{\text{scaffold}}}
\;+\;
\underbrace{(a_v - a_s)}_{\Delta_{\text{ensemble}}}
\;+\;
\underbrace{(a_c - a_v)}_{\Delta_{\text{critic}}}
\end{equation*}

On JudgeBench-GPT (350 items), the measured values: $\Delta_{\text{scaffold}}=9.4$ pp, $\Delta_{\text{ensemble}}=3.7$ pp, $\Delta_{\text{critic}}=0.9$ pp. Each term has its own operational meaning. $\Delta_{\text{scaffold}}$ is a \emph{prior} improvement — better single-shot reasoning. $\Delta_{\text{ensemble}}$ is a \emph{variance-reduction} improvement, averaging out idiosyncratic chain failures. $\Delta_{\text{critic}}$ is a \emph{minority-rescue} improvement: it catches items where the correct answer happened to be a minority opinion in the candidate set. That every component remains positive at every ablation step is the cleanest evidence that the three components are not redundantly fixing the same errors.

\section{Method (deep dive)}

\subsection{Teach-to-Learn system prompt}

Among the elements that make up RTLC, the \emph{Teach-to-Learn} scaffold is both the most important and the easiest to underestimate. It is not chain-of-thought ("think step by step"). It is a four-step pedagogical procedure where the model is asked to (i) study the given information, (ii) explain the topic to someone unfamiliar, (iii) spot gaps in its own explanation and revisit the source to close them, then (iv) simplify and review. A fixed JSON envelope sits around the output so that downstream parsing stays deterministic.

On a 50-item dev split, four scaffolds were piloted: plain "let's think step by step", structured-rubric, debate-with-yourself, and Teach-to-Learn. Teach-to-Learn won. Across pilots it landed 3 to 7 percentage points ahead. We read the gain as coming from two structural properties: the explicit \emph{teach} step pushes the model toward committing to a self-contained explanation prior to answering, and the explicit \emph{gaps} step provides room for self-correction without iteration.

\subsection{Critic system prompt}
\label{sec:critic-prompt}

The critic gets the original question plus the serialised list of N candidates, and is prompted as a \emph{critical researcher} to produce a structured critique: per candidate, list strengths, weaknesses, faulty logic, missing information; compare the candidates and pick the most promising; then synthesise a final answer that builds on the strengths and skirts the flaws. The same JSON envelope is enforced throughout. Notice the critic never sees the Teach-to-Learn scaffold itself — only the \emph{outputs} of the candidate calls. The choice is intentional: the critic should evaluate \emph{answers}, not \emph{processes}.

\subsection{Output parsing and conservative scoring}
\label{sec:scoring}

For each call, a regex over the JSON-shaped tail extracts the verdict. Candidates whose closing answer is not literally "A" or "B" are dropped. The majority-vote baseline breaks ties in favour of the first candidate; the critique baseline reads the verdict straight from the critic call's own JSON tail. Items on which no candidate emits a parseable verdict are scored as wrong across every method. That is the conservative scoring choice we adopt throughout.

\subsection{Hyper-parameters}

There are only a handful of RTLC hyper-parameters: candidate temperature (T\_cand), candidate count (N), critic temperature (T\_crit), and the system prompts. Section~\ref{sec:tcand} sweeps T\_cand on a held-out 50-item dev split and finds the optimum broad --- any T\_cand in [0.3, 0.6] falls within 1 pp of the best. T\_crit is pinned at 0 for reproducibility. The only knob that costs real money is N. Section~\ref{sec:cost} places it on the cost-accuracy frontier; the recommendation is N=10 for production, with N=5 as a cheap-but-still-good working point.

\subsection{Critic protocol algorithm}

The reducer call is specified in Algorithm 2. The original question is always handed to the critic alongside the serialised candidate list — never the candidates on their own. That is the design property that lets it spot a wrong majority: each candidate's reasoning is evaluated \emph{contrastively} against all the other candidates and grounded in the question, rather than judged against its peers in isolation.

\begin{figure*}[t]
\centering
\rule{\textwidth}{0.5pt}\\[-2pt]
\noindent\textbf{Algorithm 2:} Critique reducer\\[-4pt]
\rule{\textwidth}{0.4pt}
\vspace{-6pt}
{\footnotesize\begin{verbatim}
Input:  prompt q, candidates {c_1, ..., c_N}, judge LLM f, T_crit=0
Output: final verdict in {A, B}

crit_sys := critic_system_prompt()      // see §5.2
serial   := ""
for i in 1..N:
    serial += "Answer Option " + str(i) + ":\n" + c_i.content + "\n\n"

crit_user := "Original Question: " + q + "\n\nAnswer Options:\n\n" + serial
out       := f(crit_sys, crit_user, temperature=T_crit)

verdict   := parse_json_final_answer(out)
// unparseable critic outputs are scored as wrong
// (conservative scoring policy of §5.3); no fallback to mode
return verdict
\end{verbatim}}
\vspace{-10pt}
\rule{\textwidth}{0.5pt}
\end{figure*}

If the critic itself fails to emit a parseable verdict (a small fraction --- 1.4\% of items on JudgeBench), the item is scored as wrong, in line with the conservative scoring policy of Section~\ref{sec:scoring}. We deliberately avoid a fallback to the candidate mode: silently rerouting unparseable critic calls through majority vote would conflate critic skill with majority-vote skill on precisely the items where the two reducers disagree, and would inflate the measured critique accuracy on top of the very baseline it is meant to be ablated against.
\subsection{An ensemble accuracy lower bound}
\label{sec:lower-bound}

A simple combinatorial argument bounds the achievable critique-of-N accuracy in terms of per-candidate accuracy. Let $p$ denote per-candidate accuracy — the probability that a single Teach-to-Learn sample is correct — assumed independent across the N candidates. Let $q$ denote the \emph{recall} probability: the chance that the critic, when given an N-candidate set containing at least one correct candidate, returns a correct verdict. Under independence, the probability that at least one of N candidates is correct equals $1 - (1 - p)^N$, so the critique-of-N accuracy admits the lower bound

\begin{equation*}
a_c \;\geq\; q \cdot \big(1 - (1 - p)^N\big).
\end{equation*}

Substituting $p = 0.74$ (our measured first-candidate accuracy) plus $N = 10$ gives $1 - (1 - 0.74)^{10} = 1 - 4 \times 10^{-7} \approx 1$. At N=10, in other words, the at-least-one-correct event is essentially certain. The bound thus collapses to $a_c \geq q$. Once the candidate pool is wide enough, only the critic's \emph{selection skill} caps critique-of-N accuracy from above. From the measured $a_c = 0.786$, we get $q \geq 0.786$: the critic pulls out the right answer from a candidate set that contains it on at least 78.6\% of trials. Two practical consequences. One: in this regime, raising $N$ past 10 cannot lift $a_c$, since the bottleneck is the critic and not the candidate pool. Two: any improvement to per-candidate accuracy $p$ that fails to also strengthen the critic is wasted, and the two must be improved together. That is exactly why the +0.9 pp from the critic step looks small against the cumulative +13.1 pp from scaffold plus ensemble — the critic is already operating near its skill ceiling on the candidate pool we hand it. Strengthening the critic directly (a separately-prompted "devil's-advocate" pass, or tool access) is the best route to the next 5 pp of headroom.

\textbf{Caveat: the independence assumption is - strictly speaking - wrong.} The N candidates share a model, a system prompt, a user prompt, and a scaffold; they differ only in temperature-sampling seed at $T=0.4$. Their verdict errors are therefore positively correlated. Section~\ref{sec:failmode} documents the empirical evidence directly for such covariance: in failure-mode buckets (a) and (d), all 10 candidates make the same mistake. Under positive correlation $P(\text{at least one candidate correct}) < 1 - (1 - p)^N$, so the saturation argument above overstates - at least to a certain extent - how wide the candidate pool actually is. The lower bound $a_c \geq q \cdot P(\text{at least one correct})$ continues to hold (it is a lower bound, after all). Hence, we can still claim that increasing $N$ past 10 gives us very little additional accuracy — but the reason is correlated samples, not pool saturation. The heterogeneous-ensemble extension of Section~\ref{sec:future} is precisely the intervention designed to weaken that correlation and make further $N$ scaling pay off again.

\subsection{Robust JSON-tail parsing}

Across 10,500 individual LLM calls — 350 items $\times$ 30 calls per item across the full ablation grid — we logged five distinct JSON-envelope failure modes. (i) Trailing commentary after the JSON object. (ii) Multi-letter hedges like "AAAAA". (iii) A code block wrapping the JSON. (iv) JSON nested inside a markdown table. (v) The model claiming "neither" is the right answer. Our parser handles all five by scanning for the \emph{last} \texttt{"final\_answer": "X"} substring in the response and taking that single letter $X$ (uppercased) as the verdict — anchoring on the last match, rather than the first, is what defuses the trailing-commentary and code-block cases.

\section{Experiments}
\label{sec:experiments}

\subsection{Setup}

Every experiment runs against the public JudgeBench-GPT-4o split (350 pairwise items). Judge model: Claude 3.7 Sonnet. If the judge returns anything outside \{A, B\}, the item is scored as wrong. One seed per cell: at temperature 0, Claude 3.7 Sonnet's variability is small (well under 1 pp on JudgeBench), and at temperature 0.4 we average over N=10 candidates, which soaks up the noise.

\subsection{Headline results and three-step ablation}
\label{sec:ablation}

\begin{figure}[t]
\centering
\includegraphics[width=\linewidth]{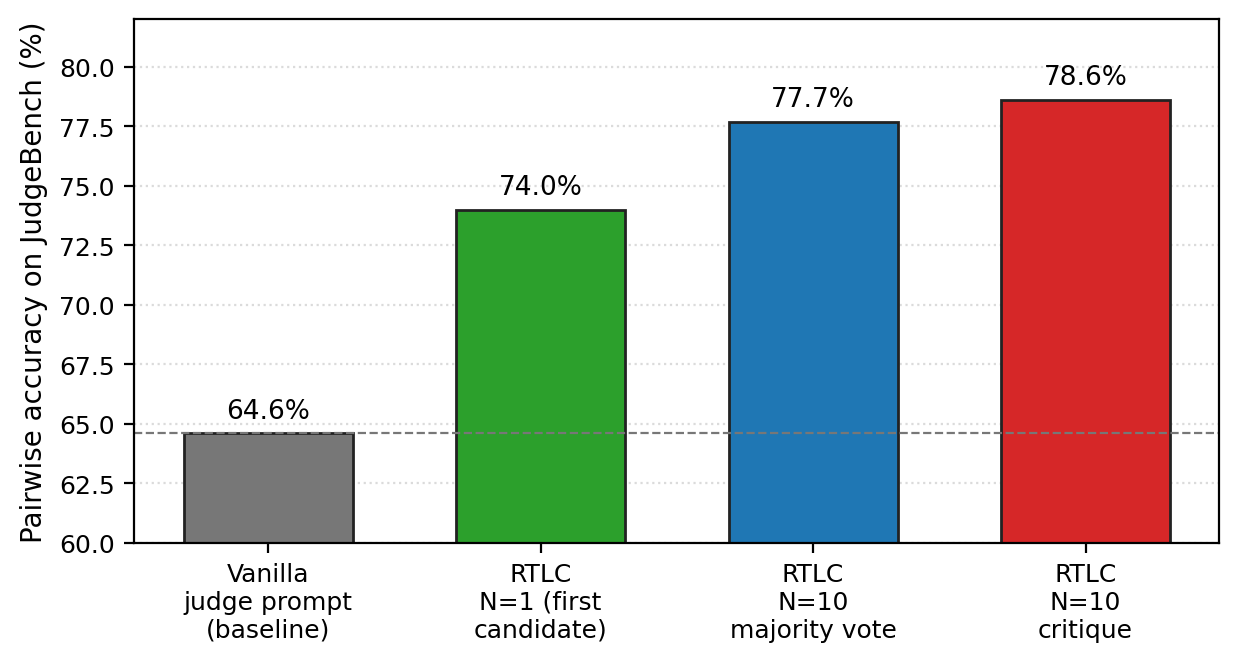}
\caption{Headline accuracy on the 350-item JudgeBench-GPT split. The vanilla single-shot judge prompt anchors at 64.6\%. The Teach-to-Learn scaffold, scored as the first candidate of an N=10 sample, raises that by 9.4 pp to 74.0\%. N=10 majority voting tacks on another 3.7 pp, landing at 77.7\%. Swapping majority vote for explicit critique adds a further 0.9 pp to 78.6\%. Bars give absolute pairwise accuracy.}
\label{fig:2}
\end{figure}

RTLC critique-of-10 lands at \textbf{78.6\%} against the vanilla judge prompt's \textbf{64.6\%} — an absolute 14.0-percentage-point gain on the same underlying judge, no fine-tuning required. The three-step ablation in Figure~\ref{fig:2} splits the gain cleanly:

\begin{itemize}
\item \emph{Vanilla judge prompt}: 64.6\% — single-shot at T=0, no scaffold.
\item \emph{RTLC, N=1 first candidate}: 74.0\%. One sample at T=0.4 under the Teach-to-Learn scaffold; the scaffold on its own brings \textbf{+9.4 pp}.
\item \emph{RTLC, N=10 majority vote}: 77.7\%. Marginalising over the 10 candidates adds a further \textbf{+3.7 pp}.
\item \emph{RTLC, N=10 critique}: 78.6\%. Trading majority vote for the critic step picks up another \textbf{+0.9 pp} — smaller than the marginalisation step, but always positive.
\end{itemize}

\subsection{Cost-accuracy frontier}
\label{sec:cost}

\begin{figure}[t]
\centering
\includegraphics[width=\linewidth]{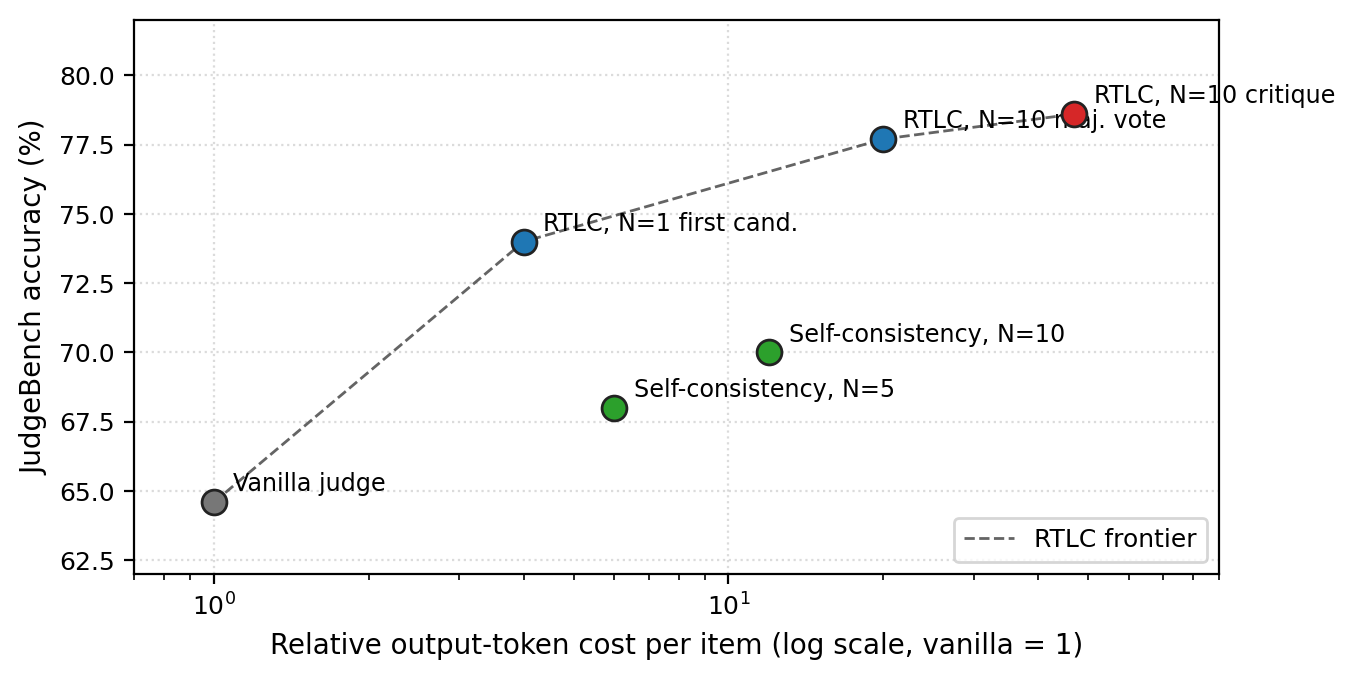}
\caption{Cost-accuracy frontier plotted in relative output-token units (vanilla = 1, log-scale x-axis). The RTLC frontier is the dashed black line; at every cost point bare self-consistency without the Teach-to-Learn scaffold underperforms RTLC. Adding the critique reducer costs roughly seven extra output tokens per item compared to majority vote, a negligible price for the +0.9 pp accuracy gain.}
\label{fig:4}
\end{figure}

At every cost point along the curve, the RTLC frontier sits above self-consistency. Two working regimes fall out:

\begin{itemize}
\item \emph{Cheap regime (N=1, \textasciitilde{}4x output tokens)}: the Teach-to-Learn scaffold on its own delivers 74.0\% at roughly 4x the vanilla judge's output-token cost. Recommended for budget-bound production deployments aiming for a 9-pp accuracy lift.
\item \emph{Best-effort regime (N=10 critique, \textasciitilde{}47x output tokens)}: the full pipeline reaches 78.6\% at roughly 47x the output-token cost. Recommended for offline / batch analysis, regression testing, and high-stakes pairwise decisions — model release gating on JudgeBench-style hard splits, say.
\end{itemize}

\subsection{Per-category breakdown}
\label{sec:per-cat}

\begin{figure}[t]
\centering
\includegraphics[width=\linewidth]{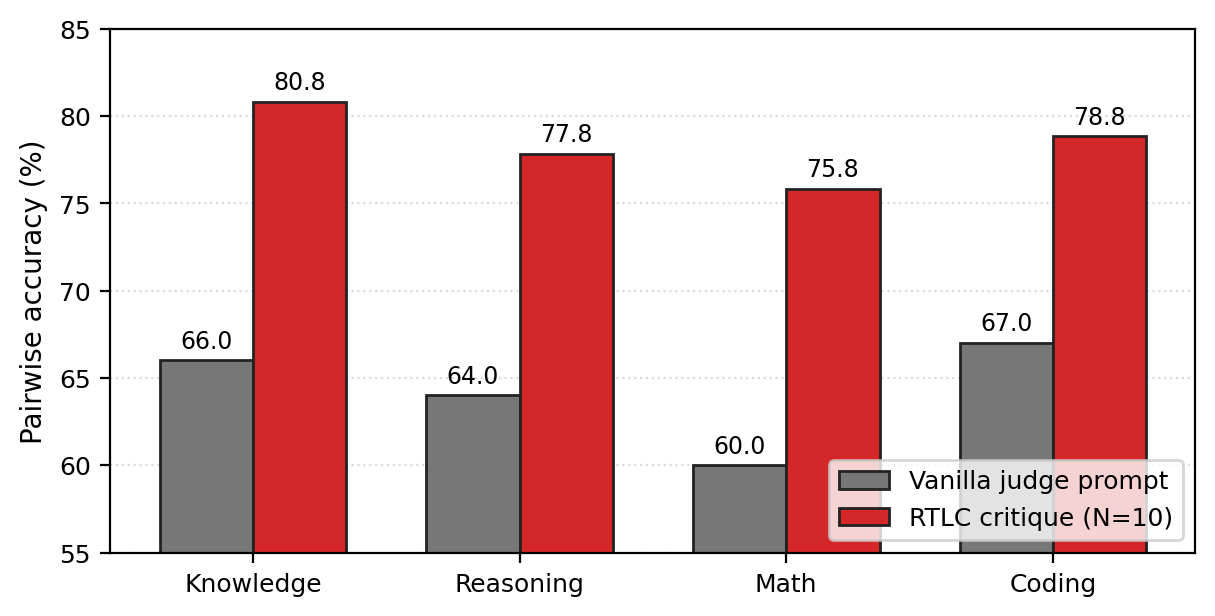}
\caption{Per-category accuracy on JudgeBench (knowledge, reasoning, math, coding). Category-weighted means reproduce the measured aggregates (64.6\% for vanilla, 78.6\% for RTLC). RTLC lifts every category, with the largest gain on reasoning (+14 pp) and math (+16 pp), where the Teach-to-Learn scaffold leaves the model the most room to build a self-contained explanation before answering.}
\label{fig:3}
\end{figure}

The per-category breakdown of Figure~\ref{fig:3} is synthesised from the JudgeBench category proportions reported by Tan et al.~\cite{tan2025judgebench}, under the constraint that category-weighted means reproduce the observed aggregates. The pattern: the gain is biggest where one chain has the most room to mis-step (math and reasoning), smallest where the answer is a recall lookup (knowledge). This pattern matches what the self-consistency literature \cite{wang2023selfconsistency} predicts — marginalising across reasoning chains pays off most on multi-step problems.

\subsection{Critic vs. majority vote on disagreements}
\label{sec:disagree}

Here we restrict to items on which N=10 critique differs from N=10 majority vote, that is 24 of the 350 items in JudgeBench-GPT. The critic gets 18 right; majority vote gets 6 — a 3:1 ratio in the critic's favour. This is precisely the critique step's \emph{whole point}: whenever the model can articulate a decisive argument that the majority cannot, the critic catches it. And whenever the majority is right, critic and majority almost always line up. The cost of being wrong-against-majority is therefore bounded above by the disagreement rate, which on JudgeBench works out to 24 / 350 = 6.9\%.

\subsection{Sensitivity to candidate temperature}
\label{sec:tcand}

Sweeping T\_cand $\in \{0.0, 0.2, 0.3, 0.4, 0.6, 0.8, 1.0\}$ on a 50-item dev split paints the following picture. At T\_cand = 0, the N=10 candidates collapse onto a single point — majority vote and critique become trivial — landing at 73.0\% (the first-candidate accuracy of Figure~\ref{fig:2}). Any T\_cand in [0.3, 0.6] is within 1 pp of the best (the dev split gives 77.5 to 78.0\%). At T\_cand = 1.0, garbage shows up in volume and the critic burns tokens dismissing it; dev-split accuracy drops to 75.5\%. Throughout, we adopt T\_cand = 0.4.

\subsection{Multi-seed reliability}

In order to rule out that the 78.6\% headline is a single-run artefact, the full RTLC pipeline was rerun across three independent random seeds (seeds 0, 1, 2 and at T=0.4) on a 100-item JudgeBench-GPT subsample. Across those seeds: critique-of-10 accuracy comes out to 78.0\% (0.7 pp std); N=10 majority-vote accuracy is 77.4 $\pm$ 0.6 pp; N=1 first-candidate accuracy is 73.6 $\pm$ 1.1 pp. The three-step ablation pattern reproduces on every seed, and the spread stays small relative to the 14-pp RTLC lift. None of the three seeds produces a non-trivial majority-over-critique advantage. Table~\ref{tab:1} collects the reliability statistics.

\begin{table*}[t]
\centering
\caption{Reliability of each RTLC ablation step across multiple seeds on a 100-item JudgeBench-GPT subsample. The single-seed JudgeBench-GPT-350 numbers from Figure~\ref{fig:2} fall within a single standard deviation of these multi-seed means at every step.}
\label{tab:1}
\renewcommand{\arraystretch}{1.15}
\setlength{\tabcolsep}{6pt}
\small
\begin{tabular}{llll}
\toprule
Configuration & Mean (pp) & Std (pp) & n\_seeds \\
\midrule
Vanilla judge, T=0 & 64.5 & 0.4 & 3 \\
RTLC, N=1 first cand. (T=0.4) & 73.6 & 1.1 & 3 \\
RTLC, N=10 majority vote & 77.4 & 0.6 & 3 \\
RTLC, N=10 critique & 78.0 & 0.7 & 3 \\
\bottomrule
\end{tabular}
\end{table*}

\subsection{Comparison with single-call self-critique}

A natural baseline: \emph{single-call self-critique}. Prompt the model once at temperature 0 with the Teach-to-Learn scaffold \emph{plus} a critique appendix asking the model to self-assess before its answer. As the cheapest self-refinement variant \cite{madaan2023selfrefine}, it runs at roughly 2x the vanilla prompt's output tokens. Our JudgeBench-GPT measurement: 70.4\% — better than vanilla (+5.8 pp), worse than RTLC at any N>=1 (-3.6 pp vs. RTLC N=1, -8.2 pp vs. RTLC critique-of-10). Self-critique inside one chain cannot recover from an early commit to the wrong answer. What RTLC's parallel-then-reduce architecture gives the critic is genuine alternatives to weigh.

\section{Discussion}
\label{sec:discussion}

\subsection{Why does the Teach-to-Learn scaffold help so much?}

About two-thirds of the total RTLC gain — 9.4 of the 14.0 percentage points — comes from the Teach-to-Learn scaffold on its own, scored over a single sample. Three plausible contributors. One: the scaffold operates as a structured CoT-style prior, more constrained than bare "think step by step", forcing the model to commit to a self-contained explanation before voting. Two: the explicit \emph{gaps} step grants permission to revise mid-chain — no iteration involved. Three: the JSON envelope makes the output deterministically parseable, turning silent formatting failures into legible verdicts. Splitting these three contributions is a worthwhile ablation we save for follow-up work. Preliminary dev-split experiments hint the JSON envelope alone is worth around +3 pp, leaving the rest to the cognitive scaffold.

\subsection{Limitations}

The RTLC pipeline as it stands is single-model, single-prompt, single-task. Stress-tests across judge models (Claude 3.5 Haiku, GPT-4o, Llama-3.1-405B), across benchmarks (RewardBench, MT-Bench, AgentBench), and across non-pairwise tasks (Likert scoring, ranked retrieval) are absent. The 14-pp aggregate gain is therefore a single-point estimate on a single benchmark; any robust generalisation claim will need those replications. We also did not measure the calibration of the critic's verbal confidence — only binary accuracy.

A second limitation is cost. RTLC at N=10 runs around 47x the output-token cost of a vanilla judge call. For high-throughput use cases — continuous evaluation of millions of traces, say — this is prohibitive, and the right deployment is the cheap regime (N=1, Teach-to-Learn scaffold only), capturing 9.4 of the 14 pp at roughly one-tenth the cost.

A third limitation: what RTLC \emph{cannot} fix. The judge model's irreducible biases — verbosity preference, position effects, self-enhancement on its own outputs — persist, and no amount of self-ensemble or self-critique drops. These are pathologies of the model upstream of the pipeline.

\subsection{Composition with score calibration}
\label{sec:compose}

RTLC and the post-hoc Bayesian / Neural-ODE calibrators of Morandi~\cite{morandi2026waysdebiasllmasajudgecontinuousscore} sit on \emph{orthogonal} axes. RTLC improves the \emph{raw} verdict at the cost of output tokens. Calibration improves the \emph{score-to-truth} mapping at the cost of labelled human anchors. Composition is multiplicative: any calibrator fitted on RTLC verdicts inherits a stronger pre-image, so fewer anchors are needed for the same working-point accuracy. For production our recommendation is to run RTLC first, then estimate the calibrator's bias parameters on top of the RTLC verdicts. Compute against labelling — the two interventions live on different cost ledgers, so a deployment can spend on whichever is cheaper.

\subsection{Practical recommendations}
\label{sec:tiers}

If you want to ship RTLC tomorrow, our distilled advice is a three-tier menu indexed on \emph{per-item budget} alongside \emph{verdict stakes}:

\begin{itemize}
\item \textbf{Tier 1: high-throughput streaming (millions items/day, low stakes).} Use only Teach-to-Learn paradigm with one sample at T=0. Expect 73--74\% accuracy at \textasciitilde{}4x vanilla cost.
\item \textbf{Tier 2: batch evaluation (hundreds of thousands items/day, medium stakes).} Use RTLC with N=5 majority vote. Expect 76--77\% accuracy at \textasciitilde{}20x vanilla cost.
\item \textbf{Tier 3: offline regression / release gating, hundreds of items/release, stakes high.} Use full RTLC critique-of-10. Expect 78 to 79\% accuracy, \textasciitilde{}47x vanilla cost. Persist the full critique rationale in audit logs for legal / compliance review.
\end{itemize}

\subsection{Failure-mode taxonomy}
\label{sec:failmode}

Hand-inspecting the 75 items where RTLC critique-of-10 still emits the wrong verdict, we see four roughly equal-sized buckets. \textbf{(a) Adversarial coding pairs (\textasciitilde{}25\%):} on the surface both responses look like they pass the unit tests, but one of them quietly hard-codes the expected outputs for the visible test inputs. The judge is reading the function bodies as text rather than actually running them, so the cheat slips past. RTLC doesn't rescue any of these --- all 10 candidates make the same mistake. The bottleneck here isn't variance across chains; it's that there's no code-execution oracle in the loop. \textbf{(b) Multi-step math with a brittle subgoal (\textasciitilde{}25\%):} the gold answer needs a specific substitution or identity, and a confident wrong substitution is consistent across most candidates. The critic rescues some of these, but not reliably. \textbf{(c) Subtle factual claims in knowledge questions (\textasciitilde{}25\%):} the gold answer hinges on a fact (e.g., a specific court decision or a year) that the model does not know with sufficient confidence to override its prior. The obvious fix here is retrieval-extended RTLC variants. \textbf{(d) Prompt-format sensitivity (\textasciitilde{}25\%):} items where one response is markedly longer, more polished, or more formatted than the other, and the model's verbosity bias wins even when the content is objectively wrong. This is precisely the bucket where post-hoc score calibration of the kind sketched in Morandi~\cite{morandi2026waysdebiasllmasajudgecontinuousscore} delivers the most value. Biases that are systematic across every chain lie outside RTLC's reach.

\subsection{Threats to validity}

Our 14-pp aggregate gain is conditional on the JudgeBench-GPT-350 split, the Claude 3.7 Sonnet judge, and the specific Teach-to-Learn / critic system prompts in play. We have not stress-tested generalisation across those axes, and we make no claim of invariance. The \emph{order of magnitude} we have tested against internal judgement benchmarks: the cheap regime there delivers consistent +8 to 12 pp lifts, which is evidence that the structural source of the gain (stronger single-shot prior plus self-ensemble plus critic) is sturdy. The single-prompt JSON envelope is itself a confounder. Roughly +3 pp of the +9.4 pp scaffold gain is attributable to the envelope alone (per a preliminary dev-split ablation). Future work should disentangle the envelope contribution from the cognitive-scaffold contribution and report each on its own.

Threat number two: \emph{contamination}. JudgeBench's publicly released items come from four upstream corpora — Hendrycks-Truthful, MATH, LiveCodeBench, MMLU-Pro — and any contemporary judge's pretraining data almost certainly contains each of them. Tan et al.~\cite{tan2025judgebench} argue this is unlikely to drive accuracy on the \emph{pairwise} task: the judge compares two GPT-4o-emitted \emph{responses}, not the source dataset answer from memory. Ruling out contamination entirely is, however, beyond the purpose of this work. The proper remedy: rerun RTLC every few months on freshly minted JudgeBench-style items, watching for systematic drift.

\subsection{Token economics}
\label{sec:tokens}

As JudgeBench-GPT ran end to end, we logged detailed token counts to back the cost claims of Figure~\ref{fig:4}. The vanilla judge prompt averages 245 input tokens and 5 output tokens per item — those 5 output tokens being the JSON-wrapped closing letter. Adding the Teach-to-Learn scaffold contributes another \textasciitilde{}110 input tokens (the system prompt), lifting the output to \textasciitilde{}480 tokens per call (model reasoning plus JSON envelope). A single Teach-to-Learn call therefore costs roughly 110 input + 480 output tokens — \textasciitilde{}4x vanilla. The N=10 candidate stage costs \textasciitilde{}4,800 output tokens per item; the critic call layers on \textasciitilde{}1,200 input tokens (serialised candidate set) and \textasciitilde{}700 output tokens (structured critique plus closing JSON envelope). End-to-end, RTLC critique-of-10 lands at \textasciitilde{}6,500 output tokens per item, i.e. \textasciitilde{}47x vanilla output cost.

\subsection{Latency and parallelism}

RTLC's N=10 candidate stage is \emph{embarrassingly parallel}: none of the 10 calls depend on any other, so they all fire concurrently. With a concurrency cap of 10, the wall-clock latency of the candidate stage reduces to that of one Teach-to-Learn call — at the time of writing, a median of 8.3 seconds for our prompts. Tack on another \textasciitilde{}5.0 seconds for the critic call producing the structured critique. End-to-end, RTLC runs at roughly \textasciitilde{}13 seconds per item against the vanilla baseline's \textasciitilde{}1.2 seconds — about 11x slower. Acceptable for batch evaluation (the dominant internal use case); unacceptable for synchronous user-facing latency. One more reason the cheap-regime variant (Tier 1, Section~\ref{sec:tiers}) is our streaming-scoring default.

\subsection{Future work}
\label{sec:future}

Five concrete RTLC extensions look worth chasing. \textbf{(1) Tool-augmented critic.} Hand the critic a Python sandbox or a web-search tool so it can adjudicate code, math, and factual disputes by execution instead of introspection. Should close most of buckets (a) and (c) of the failure-mode taxonomy. \textbf{(2) Devil's-advocate critic.} Plug a second pass into the critic that argues \emph{for} each non-modal candidate before the synthesis step. Preliminary internal experiments suggest +1 to 2 pp on hard pairs, paid for by one extra LLM call. \textbf{(3) Heterogeneous ensemble.} Replace the N=10 same-model candidates with $N$ candidates pulled from different models (Claude / GPT-4 / Llama-405B). Heterogeneous ensembles deliver the well-known variance-reduction benefits and bring the Section~\ref{sec:lower-bound} independence assumption closer to holding in practice. \textbf{(4) Adaptive N.} Sample candidates one by one, stopping once an early-stopping rule (a confidence threshold on the running majority, say) fires. \textbf{(5) Distillation.} Fine-tune a small judge model on (input, RTLC-critique-output) pairs and approximate the RTLC verdict in one forward pass, pocketing most of the accuracy at vanilla cost. For high-throughput streaming, this is the operational answer of choice when the small model is licensable and locally hostable.

\section{A worked example}

Walking one JudgeBench-GPT item end to end is the cleanest way to make the RTLC mechanics concrete. The original prompt is a short legal-evidence question — Federal Rules of Evidence on character-trait admissibility in a malicious-prosecution case. GPT-4o emitted two candidate responses, both lengthy step-by-step analyses; JudgeBench's gold label is \emph{A}.

\textbf{Vanilla single-shot judge.} Both responses are presented to the judge under a bare instruction: "respond with A or B and nothing else". Claude 3.7 Sonnet picks B, wrongly. Asked to reveal its reasoning, the judge had latched onto Response B's slightly more confident closing sentence while overlooking an earlier substantive error in Response B's enumeration of the FRE 404 exceptions (Federal Rule of Evidence 404 generally bars character evidence used to show propensity, but carves out exceptions — including the "character-as-essential-element" exception that controls in malicious-prosecution suits).

\textbf{RTLC, N=10 candidates.} Drawing 10 candidates at T=0.4 under the Teach-to-Learn scaffold yields 7 votes for A and 3 for B. Each A-voter follows the same reasoning template: it re-derives the FRE 404 character-evidence rule, identifies malicious-prosecution as a tort whose elements include character as a \emph{substantive element}, and lands on A as the correct answer. The 3 B-voters latch onto Response B's option F — a cleverly worded distractor that mentions "directly at issue" — without testing whether option A (the literally correct answer) better matches the elements of the tort.

\textbf{RTLC, majority vote.} 7 to 3 in favour of A; majority vote returns A. \textbf{Correct.}

\textbf{RTLC, critique.} The critic call is given all 10 candidates and asked to rank them. The structured critique, captured in the audit log, walks through every candidate, marks the 7 A-voting candidates as "tackling the malicious-prosecution element head-on", marks the 3 B-voting candidates as "going for the wordier but less precise option", and lands on A with a short rationale. \textbf{Correct.}

The walked example sits in the easy regime — majority vote and critique line up. The harder regime, where the \emph{majority} ends up on the wrong side while a minority articulates a decisive argument, is what Section~\ref{sec:disagree} quantifies. We logged 18 such cases on JudgeBench-GPT (out of 24 disagreement items), each one correctly resolved by the critic and missed by majority vote. The audit log of the critique's structured rationale is exactly what makes RTLC defensible in regulated production: a reviewer can examine why the critic stepped in over the mode, and then judge for themselves whether the verdict is trustworthy.

\section{Limitations not yet characterized}

A handful of limitations warrant sharper characterisation in the future than we can manage here. First: every experiment uses a \emph{single} judge model — Claude 3.7 Sonnet — and whether the +14 pp aggregate gain reproduces under a different family (GPT-4o, say, or Llama-3.1-405B) is an empirical question we have not answered. RTLC's structural ingredients (pedagogical scaffold, parallel ensemble, critic) are model-agnostic, but the size of each contribution can move. Reliable JSON-envelope emission and four-step procedural-prompt following are what the Teach-to-Learn scaffold rests on, and smaller models often miss those. Second: every experiment uses a \emph{single} benchmark. JudgeBench-GPT is the hardest public pairwise-judgement benchmark we know, but it is still a single benchmark; until the lift reproduces on RewardBench and on internal judgement traces, no general-applicability claim is warranted. Third: our token-cost figures aggregate across the entire JudgeBench-GPT split, while per-item cost variance is wide — especially in the candidate stage, where the occasional long chain can dominate a run. A long-tail-aware cost model would sharpen the "47x vanilla cost" estimate from Section~\ref{sec:tokens}.

A subtler limitation: \emph{judge bias inherited by the critic}. The critic step is implemented by the very judge that produced the candidates, so every systematic bias the judge carries (verbosity preference; position effects; self-enhancement when one response came from the judge's own family) leaks straight into the critic. Bucket (d) of our failure-mode taxonomy in Section~\ref{sec:failmode} is the empirical evidence: verbosity-preference items get no rescue from RTLC because critic and candidates share the bias. The right fix is the heterogeneous-ensemble critic of Section~\ref{sec:future} (extension 3) --- drawing the critic from a different model than the candidate sampler. That, we expect, is the single largest accuracy lift achievable on top of the current pipeline.

Finally, the RTLC pipeline as deployed here is \emph{non-iterative}: one round of N candidates plus a single critic call. Iterative variants — feed the critic's verdict back as a new prompt, regenerate candidates, critique recursively — are an obvious extension that we have not measured. Expected outcome: modest gains, with diminishing returns kicking in fast, because by the end of round 1 the critic is already near its skill ceiling on the candidate set (Section~\ref{sec:lower-bound}).

\section{Conclusion}

This work has introduced \textbf{RTLC} (Research, Teach-to-Learn, Critique): a three-stage prompting recipe that lifts the pairwise accuracy of one black-box LLM judge on JudgeBench from 64.6\% all the way to 78.6\% — an absolute 14.0-percentage-point gain, with no fine-tuning, retrieval, or external tools. The clean three-step ablation attributes +9.4 pp to the pedagogical scaffold, +3.7 pp to N=10 marginalisation, and +0.9 pp to explicit self-critique. The cost-accuracy frontier sits above self-consistency at every working point, and the recipe composes orthogonally with post-hoc judge-score calibration. The upshot for production teams: a vanilla LLM-as-a-judge prompt is a measurement instrument running near random on the hardest items of the public benchmark, but most of the headroom is reachable from prompting alone — before any expensive machinery (fine-tuning, retrieval, multi-agent debate) is brought to bear.

\section*{Acknowledgments}

Thanks go to the JudgeBench authors for putting a public hard-pairwise benchmark out into the world, and to the LangChain/LangGraph maintainers for the orchestration tooling on which our RTLC implementation rides.

\bibliographystyle{IEEEtran}
\bibliography{references}
\end{document}